\documentclass[11pt,a4paper]{article}
\usepackage{geometry}
\usepackage{newtxtext,newtxmath} 
\usepackage{microtype}

\usepackage{setspace}          
\usepackage{amsmath}

\usepackage{algorithm}
\usepackage{algorithmic}

\usepackage{caption}

\usepackage{url}
\usepackage{graphicx}  
\usepackage{float}  

\usepackage[colorlinks=true, citecolor=blue, linkcolor=blue, urlcolor=blue]{hyperref}

\begin{document}

\title{Positive-Only Drifting Policy Optimization}

\author{Qi Zhang}
        
\date{}
\maketitle


\begin{abstract}

In the field of online reinforcement learning (RL), traditional Gaussian policies and flow-based methods are often constrained by their unimodal expressiveness, complex gradient clipping, or stringent trust-region requirements. Moreover, they all rely on post-hoc penalization of negative samples to correct erroneous actions. This paper introduces Positive-Only Drifting Policy Optimization (PODPO), a likelihood-free and gradient-clipping-free generative approach for online RL. By leveraging the drifting model, PODPO performs policy updates via advantage-weighted local contrastive drifting. Relying solely on positive-advantage samples, it elegantly steers actions toward high-return regions while exploiting the inherent local smoothness of the generative model to enable proactive error prevention. In doing so, PODPO opens a promising new pathway for generative policy learning in online settings.
\end{abstract}

\section{Introduction}

This paper introduces Positive-Only Drifting Policy Optimization (PODPO). Drawing on the core idea of Drifting Models~\cite{podpo_drifting}, PODPO reframes policy optimization as advantage-weighted drifting vector learning. Unlike existing methods, PODPO performs explicit drifting updates exclusively on positive-advantage samples. It leverages the inherent local smoothness of the generative model to achieve indirect error correction and proactive error prevention on near-boundary samples. At the same time, it suppresses extreme gradients at their source through a multi-temperature non-normalized accumulation mechanism, completely eliminating the need for explicit gradient clipping or trust-region constraints. The approach requires no policy likelihood computation and employs a lightweight contrastive drifting loss, making it fully compatible with the standard PPO framework while remaining straightforward to implement and deploy.

The main contributions of this paper are as follows:
\begin{enumerate}
\item We propose the PODPO algorithm, which seamlessly integrates Drifting Models with online reinforcement learning to enable likelihood-free, single-step generative policy optimization.
\item We design a per-observation local contrastive drifting mechanism that eliminates cross-sample interference, achieving higher drifting precision while preserving computational efficiency.
\item We show that positive-advantage samples alone are sufficient to achieve proactive error prevention in policy optimization, and we introduce a novel paradigm for generative reinforcement learning: ``optimize only the recoverable errors while ignoring irrecoverable states.
\item Through advantage weighting and gradient analysis, we demonstrate the method’s simplicity and training stability, showing that it can be directly embedded into existing PPO training pipelines.
\item We validate the effectiveness of PODPO on robotic continuous control tasks, providing a concise and highly efficient new solution for generative online reinforcement learning.
\end{enumerate}

\section{Background}

\subsection{PPO} PPO~\cite{podpo_ppo} typically optimizes a unimodal Gaussian policy, which struggles to capture the multimodal characteristics inherent in complex action spaces. Moreover, it relies on post-hoc penalization of negative samples for error correction, thereby limiting the policy’s exploration capability and expressive power in challenging scenarios.

\subsection{FPO/FPO++} FPO/FPO++~\cite{podpo_fpo, podpo_fpoplus} introduce generative models into the policy learning framework, effectively breaking through the bottleneck of unimodal Gaussian policies. However, existing flow-based strategies still fully inherit PPO’s explicit gradient clipping and complex trust-region designs, and they continue to depend on bidirectional updates from both positive and negative samples.

\subsection{Diffusion Reinforcement Learning and Offline Applications}
Diffusion models possess powerful distribution modeling capabilities and have been widely adopted in reinforcement learning. Current research on diffusion-based RL focuses primarily on offline settings. While diffusion models excel at learning complex multimodal policies from offline datasets, they require multi-step iterative generation during inference, incurring substantial computational overhead. This makes them ill-suited for the real-time demands of online reinforcement learning~\cite{podpo_diffusion_policy, podpo_diffuseloco, podpo_diffusion_multitask}.

\subsection{Drifting Models}
In 2026, the team of Kaiming He introduced Drifting Models, a novel single-step generative paradigm that fundamentally rethinks the ``iterative inference'' paradigm of traditional diffusion and flow-matching models by shifting the entire iterative process into the training phase.

The core idea of Drifting Models is to introduce a \emph{drifting field} that governs the movement direction of generated samples during training. When the generated distribution deviates from the target data distribution, the drifting field guides the samples toward the target; once the two distributions match perfectly, the drifting field becomes zero and equilibrium is reached. The training objective is remarkably simple: compute the drifting vector for each generated sample and train the network to output exactly that drifted target~\cite{podpo_drifting}.

Compared with traditional generative models, Drifting Models offer three compelling advantages:
\begin{itemize}
    \item \textbf{True single-step generation:} At inference time, only a single forward pass is required to produce high-quality samples, eliminating any iterative steps and delivering exceptional computational efficiency.
    \item \textbf{Extremely simple implementation:} Training requires nothing more than a plain MSE loss, with no need for complex time-step embeddings, ODE solvers, or adversarial training.
    \item \textbf{Inherent contrastive learning property:} The drifting field naturally performs distribution matching through an ``attract positive samples, repel negative samples'' contrastive mechanism, effectively capturing both local and global structures in the data.
\end{itemize}

Drifting Models have achieved breakthrough performance in image generation tasks. However, their application in reinforcement learning remains largely unexplored. This work builds upon this latest advance by integrating Drifting Models with online reinforcement learning, resulting in the proposed Positive-Only Drifting Policy Optimization (PODPO) algorithm.

\section{Positive-Only Drifting Policy Optimization}

\subsection{Gradient-Based Policy Optimization}
In online policy gradient optimization, PPO is typically employed to jointly optimize both the value function and the policy. The surrogate objective of standard PPO is given by:\[
\max_{\theta} \mathbb{E}_{a_t \sim \pi_{\theta_{\text{old}}}} \left[
    \min\left(
        r(\theta)\hat{A}_t,\ 
        \text{clip}\left(r(\theta), 1-\epsilon_{\text{clip}}, 1+\epsilon_{\text{clip}}\right)\hat{A}_t
    \right)
\right]
\]
where \(r(\theta)\) denotes the likelihood ratio between the new and old policies~\cite{podpo_ppo}.
\\
Recent flow-based policy gradient methods (such as FPO and FPO++) further extend this framework. FPO++ optimizes and clips gradients via per-sample ratio and the ASPO trust region~\cite{podpo_fpo, podpo_fpoplus}:
\[
\hat{\rho}^{(i)}_{\text{FPO++}}(\theta) = \exp\left(l_{\text{old}}^{(i)} - l_{\theta}^{(i)}\right)
\]
\[
\psi_{\text{ASPO}}\bigl(\hat{\rho},\hat{A}_t\bigr)=
\begin{cases}
\psi_{\text{PPO}}\bigl(\hat{\rho},\hat{A}_t\bigr) & \hat{A}_t\ge 0,\\[4pt]
\psi_{\text{SPO}}\bigl(\hat{\rho},\hat{A}_t\bigr) & \hat{A}_t<0
\end{cases}
\]
Both approaches achieve stable optimization by explicitly clipping gradients. In contrast, PODPO follows an entirely different path: it directly employs an advantage-weighted MSE drifting loss that requires no likelihood computation, yet still achieves implicit gradient control through advantage weighting, thereby maintaining high compatibility with the standard PPO training framework.

\subsection{Drifting Loss}
The theoretical foundation of PODPO is directly inspired by the recently proposed Drifting Models (generative drifting models)~\cite{podpo_drifting}. This work introduces a novel generative paradigm: during training, a \emph{drifting field} guides the pushforward distribution to gradually evolve toward the target data distribution, thereby achieving true single-step generation. Its core drifting loss is given by:
\[
L_j = \text{MSE}\left( \tilde{\phi}_j(x) - \text{sg}\left( \tilde{\phi}_j(x) + \tilde{V}_j \right) \right)
\]
where \(\tilde{V}_j\) denotes the drifting field and \(sg\) is the stop-gradient operation. This loss enables the network to directly learn a single-step ''drift-toward-the-target'' mapping during training, completely avoiding the multi-step iterative process required by diffusion models. PODPO naturally extends this elegant drifting-model paradigm to online reinforcement learning policy optimization.
\subsection{PODPO Drifting Loss}
The core innovation of PODPO lies in performing \emph{per-observation local contrastive drift} independently for each observation within a batch. For every positive-advantage sample \((\hat{A} > 0)\), we generate \(G\) dedicated candidate actions exclusively for that observation. These candidates are independently sampled from a standard Gaussian \(\mathcal{N}(0,1)\) at each step, forming a completely isolated 1-positive + \(G\)-negative contrastive set. This design eliminates cross-observation noise interference and thereby guarantees maximum drifting efficiency for each individual state.

Formally, for each positive-advantage sample \(b=1,\dots,B_{\rm pos}\) in the mini-batch:
\[
x^{(b)} \in \mathbb{R}^{G \times D},\ 
y_{\text{pos}}^{(b)} \in \mathbb{R}^{1 \times D},\ 
y_{\text{neg}}^{(b)} = x^{(b)}
\]
where \(x^{(b)}\) denotes the \(G\) candidate actions independently generated for the \(b\)-th observation, \(y_{\text{pos}}^{(b)}\) is the true rollout action of that observation, and the drifting vector \(V^{(b)}\) is computed solely from the positive-negative sample set belonging to this specific observation:
\[
V^{(b)} = \text{compute\_V}\left(x^{(b)}, y_{\text{pos}}^{(b)}, y_{\text{neg}}^{(b)}; T\right)
\]
The drifting target is constructed as:
\[
\text{target}^{(b)} = x^{(b)} + V^{(b)}
\]
The drifting loss is an advantage-weighted mean squared error, where the advantage weight is defined as \(w(\hat{A}^{(b)}) = \beta |\hat{A}^{(b)}|\):
\[
L_{\text{drift}} =
\frac{1}{B_{\rm pos}} \sum_{b=1}^{B_{\rm pos}}
w\left(\hat{A}^{(b)}\right) \cdot
\frac{1}{G} \sum_{g=1}^{G}
\left\| x_g^{(b)} - \text{target}_g^{(b)} \right\|_2^2
\]
Note that the drifting vector \(V^{(b)}\) is computed with the stop-gradient operation.
\\
In contrast to global batch drifting, where all actions across the entire batch are pooled together for contrastive computation, PODPO’s per-observation design strictly isolates the contrastive set for each individual observation. This eliminates cross-sample interference while simultaneously preserving the high efficiency of batched parallel computation.
\subsubsection{Positive Sample Selection and Advantage Normalization}

The advantage weighting for positive samples in the drifting process, \(w = \beta |\hat{A}|\), critically relies on accurate positive-negative sample partitioning. We directly adopt the advantage values produced by the PPO critic network for this partitioning: samples with \(\hat{A} > 0\) are treated as positive, while those with \(\hat{A} \le 0\) are regarded as negative and are immediately discarded. To ensure stability in the number of positive samples, we apply zero-mean, unit-variance normalization to the advantages. This keeps the proportion of positive samples relatively consistent throughout training and effectively prevents drastic fluctuations in sample counts caused by shifts in the advantage distribution.

\subsubsection{Gentle Optimization of Boundary Samples and Proactive Error Prevention}

The clipping mechanism in traditional PPO has one important limitation. It uses clipping to control how much the new policy can reduce the probability of bad actions. This helps prevent over-punishment. However, it cannot tell the difference between two very different kinds of negative samples. 

One kind is when the environment has become extremely bad and no action can save the situation. The other kind is when the agent makes a small mistake but can still fix it with a better action. Continuing to optimize the first kind actually hurts the policy. The second kind is exactly the moment that needs the most attention and repair. 

Because PPO does not take advantage of local smoothness in the state space, it cannot make such fine-grained decisions. It can only apply a simple one-size-fits-all penalty after the fact.

PODPO naturally solves this problem. It applies no drifting loss at all to extremely bad observations. As a result, it never pulls those hopeless states in the wrong direction. For positive-advantage samples, we generate \(G\) on-policy candidate actions as negative samples and let these generated actions drift toward the positive sample.

\textbf{\textit{This mechanism not only corrects recoverable errors that have already occurred. More importantly, it actively prevents future errors: as soon as any bad trend appears, the drifting field immediately pulls the policy back to high-return regions. }}

This smart behavior—``optimize only the recoverable errors and completely ignore the ones that are already beyond repair''—is exactly why PODPO is more efficient and more adaptive than PPO clipping. It allows the policy to achieve more precise and gentler optimization in complex environments.

In short, this mechanism reveals a fresh design paradigm for generative reinforcement learning: there is no need to spend heavy effort explicitly handling negative samples or fixing errors after they happen. By simply focusing on positive samples and using the natural smoothness of the generative model, we can proactively prevent errors and indirectly correct recoverable ones, leading to more efficient and stable policy learning.

\subsubsection{Comparison with Multi-Step Diffusion Methods}

FPO and FPO++ require modeling the denoising process as a multi-step Markov Decision Process (MDP), whereas PODPO directly accomplishes policy optimization through single-step generation. This elegantly avoids prolonging the credit-assignment horizon and eliminates the need for custom samplers, resulting in significantly simpler training and deployment. More importantly, FPO++ adopts a per-sample ratio mechanism (computing independent ratios and trust regions for multiple Monte-Carlo samples within each action)~\cite{podpo_fpo, podpo_fpoplus, podpo_diffusion_policy}, while PODPO employs a strict per-observation local contrastive drift (independently generating \(G\) candidate actions for each observation and computing the drifting field in complete isolation). This achieves a more thorough ``each optimizes its own'' paradigm and effectively prevents cross-observation noise interference.

\subsubsection{Final Total Loss}

By combining the drifting loss with the standard clipped value loss of PPO, we obtain the overall loss of PODPO:
\[
L = L_{\text{drift}} + L_{\text{value}}
\]

\textbf{Straightforward implementation:} PODPO is remarkably easy to implement. One simply replaces the surrogate loss in PPO with the drifting loss \(\mathcal{L}_{\rm drift}\) and adds a noise input to the actor network (to generate the \(G\) candidate actions). The entire algorithm can then be executed using the standard PPO training pipeline.

\begin{algorithm}[H]
\caption{Positive-Only Drifting Policy Optimization (PODPO)}
\begin{algorithmic}[1]
\REQUIRE Observation $\text{obs}$, Advantage $\hat{A}$, Temperature set $\tau$, Hyperparameter $\beta$
\ENSURE Trained actor network, critic network

\STATE Initialize actor network, critic network

\FOR{each iteration}
    \STATE Collect trajectories: $\text{obs}, \text{actions}, \text{rewards}$
    \STATE Compute advantage $\hat{A} \leftarrow \text{normalized advantage}$
    \STATE Filter positive advantage samples: $\text{obs}^+ \in \mathbb{R}^{B\times\cdots}, a^+ \in \mathbb{R}^{B\times D}$ where $\hat{A}>0$

    \FOR{each $\text{obs}^+$ in positive advantage samples}
        \STATE $x \leftarrow \text{actor}(\text{obs}^+; \text{noise}) \quad x \in \mathbb{R}^{(B\times G)\times D}$
        \STATE \textbf{with} $\text{stop-gradient on V}$:
        \STATE \quad $V \leftarrow \text{compute\_V}(x, a^+, x; \tau)$
        \STATE \quad $\text{target} \leftarrow x + V$
        \STATE $L_{\text{drift}} \leftarrow \beta|\hat{A}| \cdot \text{MSE}(x, \text{target})$
    \ENDFOR

    \STATE $L_{\text{value}} \leftarrow \text{clipped value loss}$
    \STATE $L \leftarrow L_{\text{drift}} + L_{\text{value}}$
    \STATE Update actor and critic via $\text{Adam}(L)$
\ENDFOR
\end{algorithmic}
\end{algorithm}

\subsubsection{Multi-Temperature Variance Compression in the Contrastive Drifting Field: Why PODPO Naturally Requires No Clipping}

PODPO inherits the multi-temperature design proposed in the original Drifting Models paper. It computes the final drifting vector through a simple non-normalized summation across temperatures:
\[
V_{\text{total}}(x) = \sum_{\tau \in T} V_\tau(x),\quad T = \{0.02, 0.15, 2.0\}
\]
This elegant design is the core mechanism that allows PODPO to operate without any PPO-style clipping.

We model the drifting vector at each temperature as a random variable \(V_\tau = \mu_\tau + \epsilon_\tau\), where \(\mu_\tau = \mathbb{E}[V_\tau]\) represents the expected drifting force at temperature \(\tau\), and \(\epsilon_\tau\) is a zero-mean noise term. The noise terms across different temperatures are approximately weakly correlated (due to the pronounced differences in softmax weights at distinct temperatures). Consequently, the expectation and variance of the total drifting vector satisfy:
\[
\mathbb{E}[V_{\text{total}}] = \sum_{\tau} \mu_\tau \approx \mu_{0.15}
\]

We define the \emph{relative variance} (RV) as the key metric for quantifying the stability of the drifting vector:
\[
\text{RV}(V) = \frac{\|\mathbb{E}[V]\|^2}{\mathbb{E}\left[\|V\|^2\right]}
\]
After non-normalized accumulation across multiple temperatures, the overall expectation is dominated by the medium-temperature term, while the total variance is simply the sum of the individual temperature variances. As a result, the relative variance of the aggregated vector is substantially lower than that of any single low-temperature case:
\[
\text{RV}(V_{\text{total}}) \ll \text{RV}(V_{0.02})
\]

This variance-compression mechanism fundamentally suppresses extreme drifting vectors, enabling PODPO to maintain highly stable training even without any explicit gradient clipping.

Although the multi-temperature loop may appear to introduce theoretical overhead, the number of particles \(G\) per observation is small (typically 8). The overall batched computational complexity remains only \(\mathcal{O}(B_{\rm pos} \cdot G^2)\). The dominant \(\texttt{cdist}\) operation is extremely fast, so the impact on training speed is negligible. Furthermore, all distance and scaling operations are performed outside the temperature loop with unified batch processing, further ensuring high efficiency.

\section{Experiments}

\subsection{Locomotion Benchmarks}

\begin{figure}[H]
  \centering
  \includegraphics[width=0.7\linewidth]{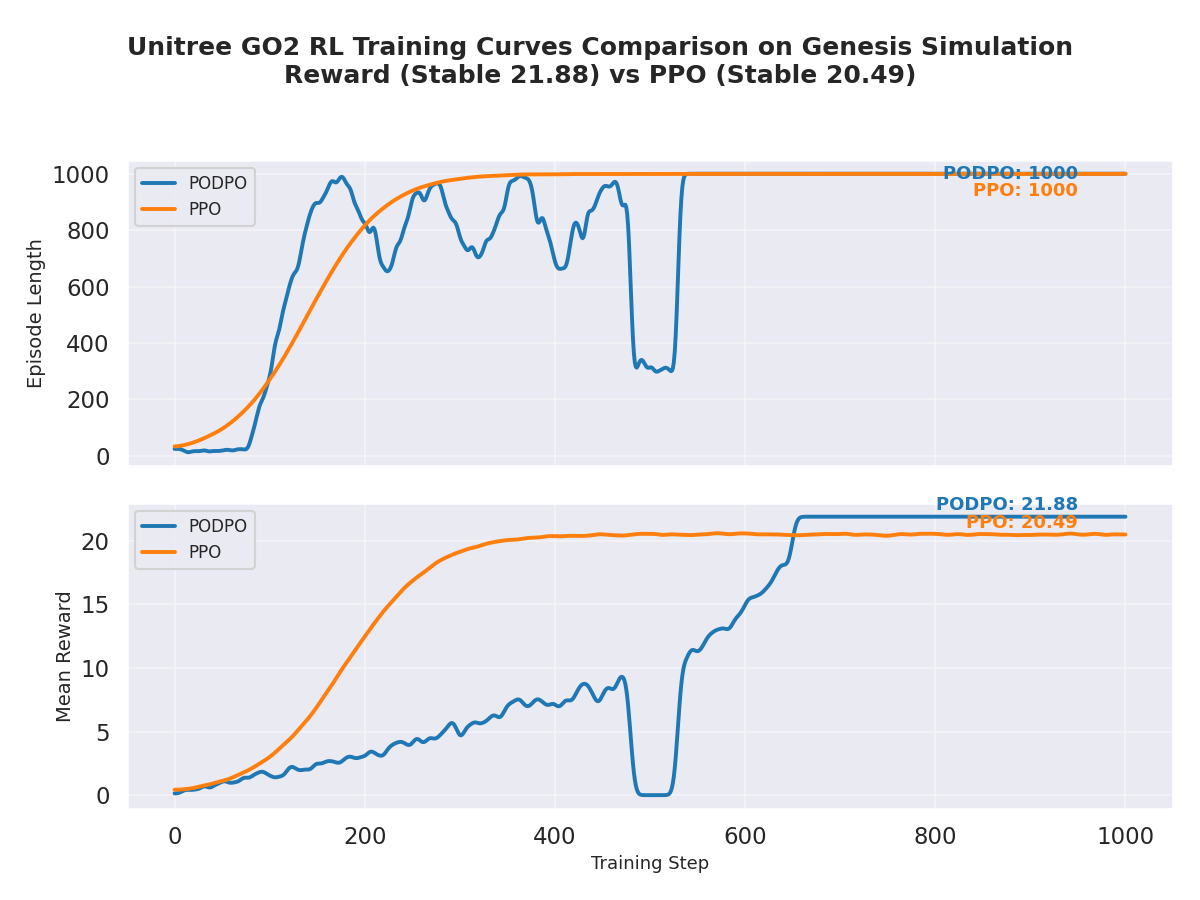}
  \caption{Comparison of PODPO and PPO on the Unitree GO2 quadruped gait locomotion task in the Genesis simulator~\cite{podpo_genesis_sim}, using identical network architecture and training configurations. The experimental results show that PODPO achieves approximately 6.7\% higher converged return than PPO, validating the superior expressiveness and exploration capability of multimodal policies in high-dimensional continuous action spaces. Although PODPO exhibits mild oscillations during the early training phase—arising from its broad exploration of the multimodal action space—once the optimal gait pattern is discovered, the policy rapidly stabilizes and consistently maintains high performance.}
  \label{fig:go2}
\end{figure}

\begin{figure}[H]
    \centering
    \includegraphics[width=1.0\linewidth]{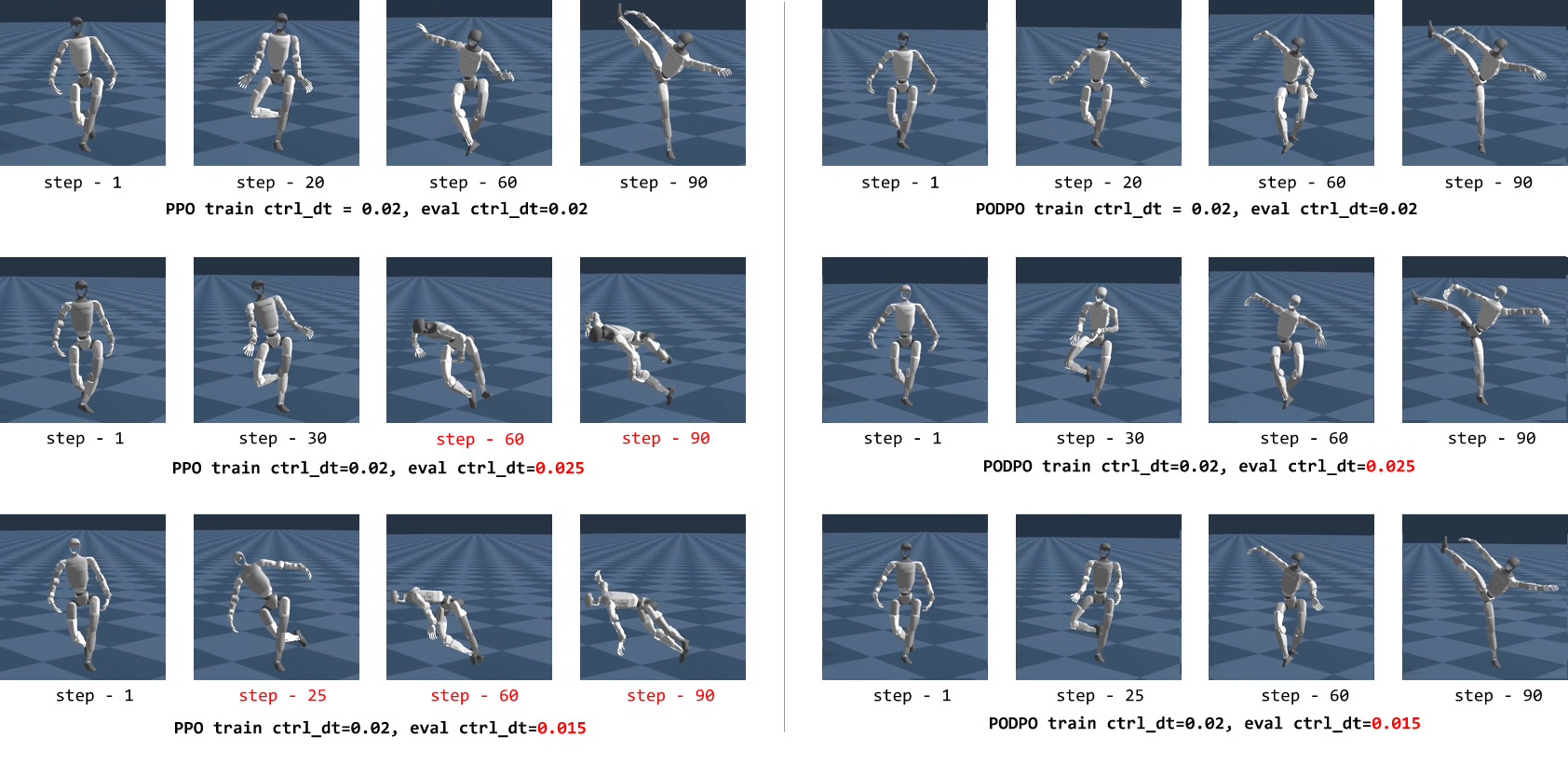}
    \caption{In the Genesis simulator~\cite{podpo_genesis_sim}, we evaluate a challenging 448-step high-difficulty dance motion tracking task under the demanding condition of no historical observations in the input (obs). PPO’s unimodal Gaussian policy is highly sensitive to control frequency and tends to collapse due to its limited single-mode action distribution. In contrast, PODPO, powered by its multimodal generative policy, adapts significantly better to dynamic environments and varying control frequencies.}
    \label{fig:placeholder}
\end{figure}

\subsection{Ablation Study}

\begin{figure}[H]
    \centering
    \includegraphics[width=1.0\linewidth]{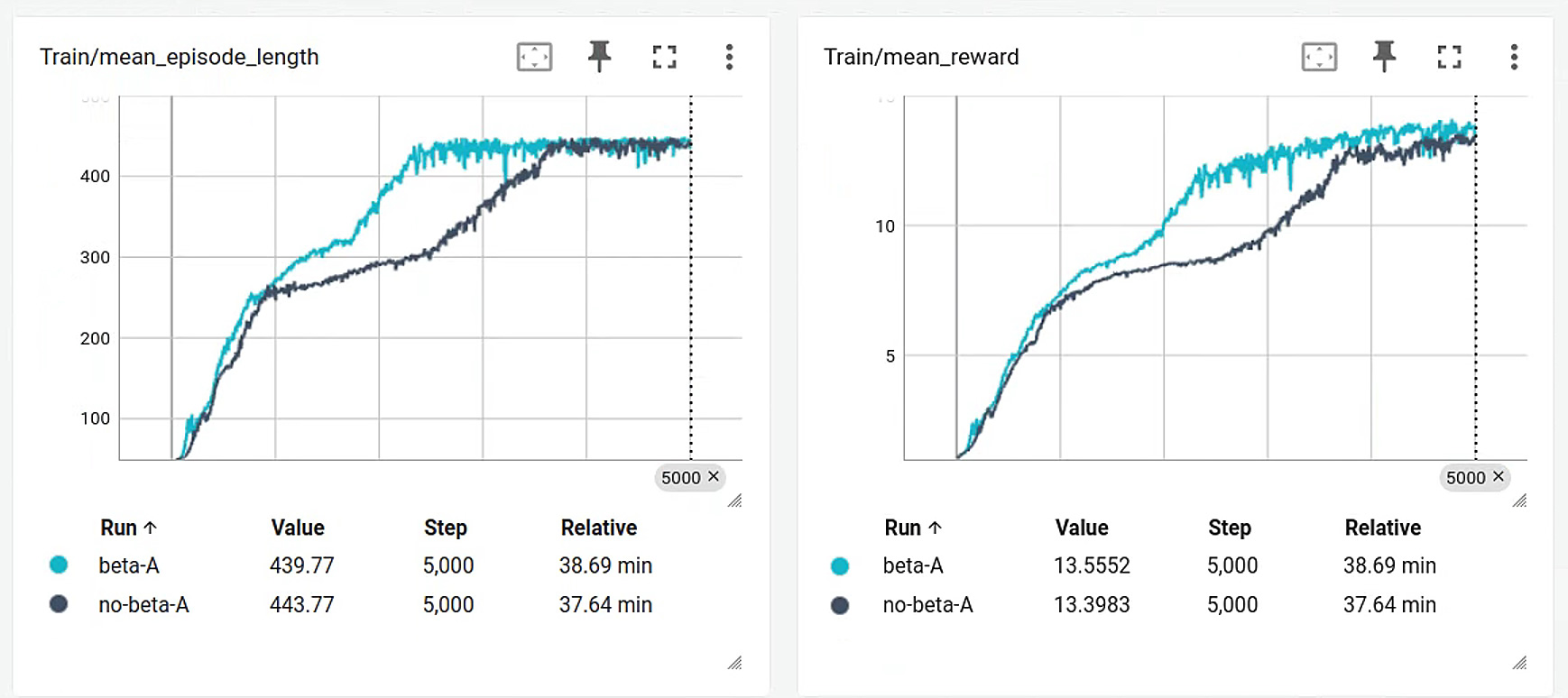}
    \caption{Ablation study on advantage weighting conducted in the Genesis simulator for the high-difficulty dance motion tracking task. The curves compare training with \(\beta|\hat{A}|\) weighting on positive samples versus training without any advantage weighting. The results clearly demonstrate that applying advantage weighting yields substantially better performance. This confirms that assigning higher weights to higher-advantage (i.e., better) samples in the early training phase effectively accelerates convergence. We adopt a simple linear weighting with \(\beta=0.1\); alternative nonlinear schemes such as \(\ln(1+x)\) or \(\tanh(x)\) could also be explored. We additionally experimented with scaling the norm \(\|V\|\). Note that, in gradient-based optimization, scaling the magnitude of \(V\) during the update is mathematically equivalent to weighting the loss term.}
    \label{fig:placeholder}
\end{figure}

\begin{figure}[H]
    \centering
    \includegraphics[width=1.0\linewidth]{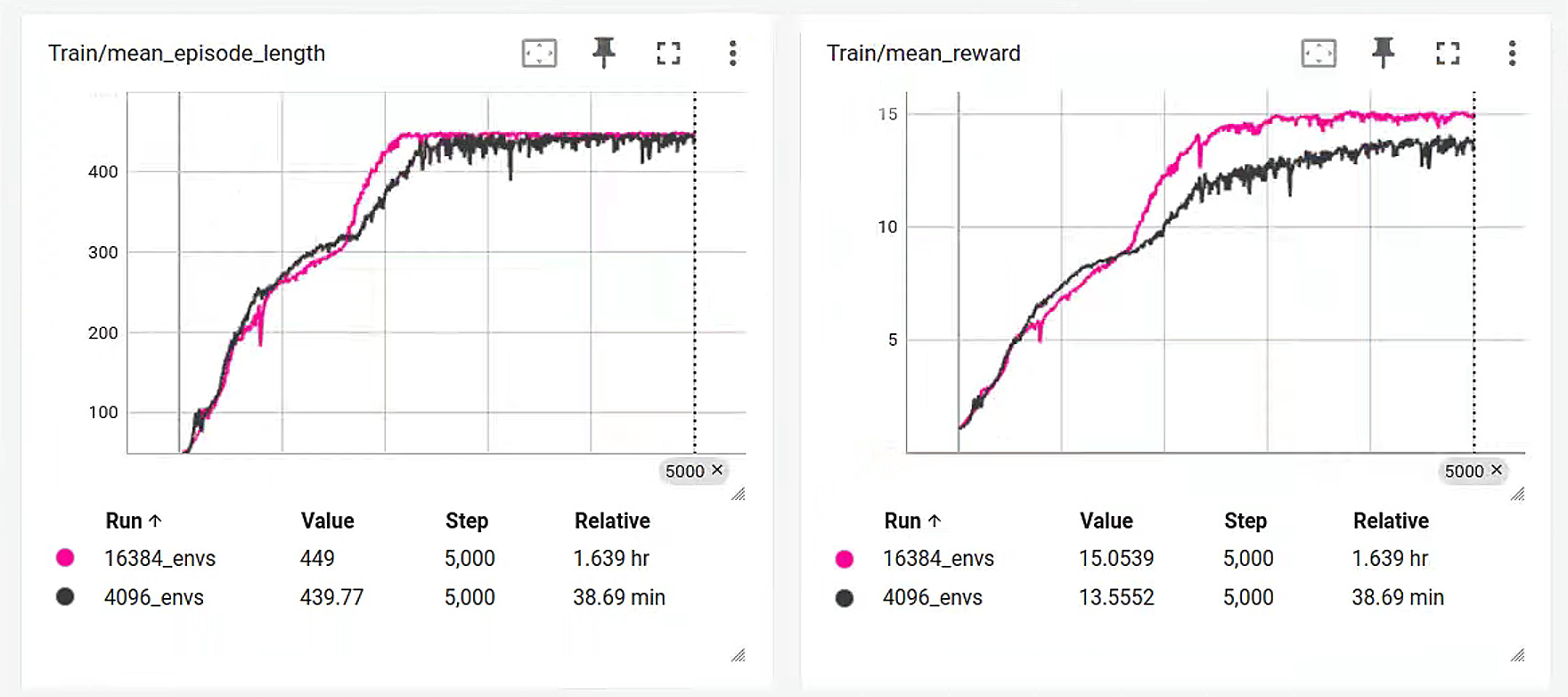}
    \caption{In the Genesis simulator, we train the high-difficulty dance motion tracking task under two different numbers of parallel environments (16384 versus 4096). The results clearly demonstrate that a higher proportion of repeated actions among the positive samples (relative to the total number of actions) leads to superior performance. This finding is consistent with PPO. However, PODPO exhibits a stronger dependence on a large number of parallel environments.}
    \label{fig:placeholder}
\end{figure}

\begin{figure}[H]
    \centering
    \includegraphics[width=1.0\linewidth]{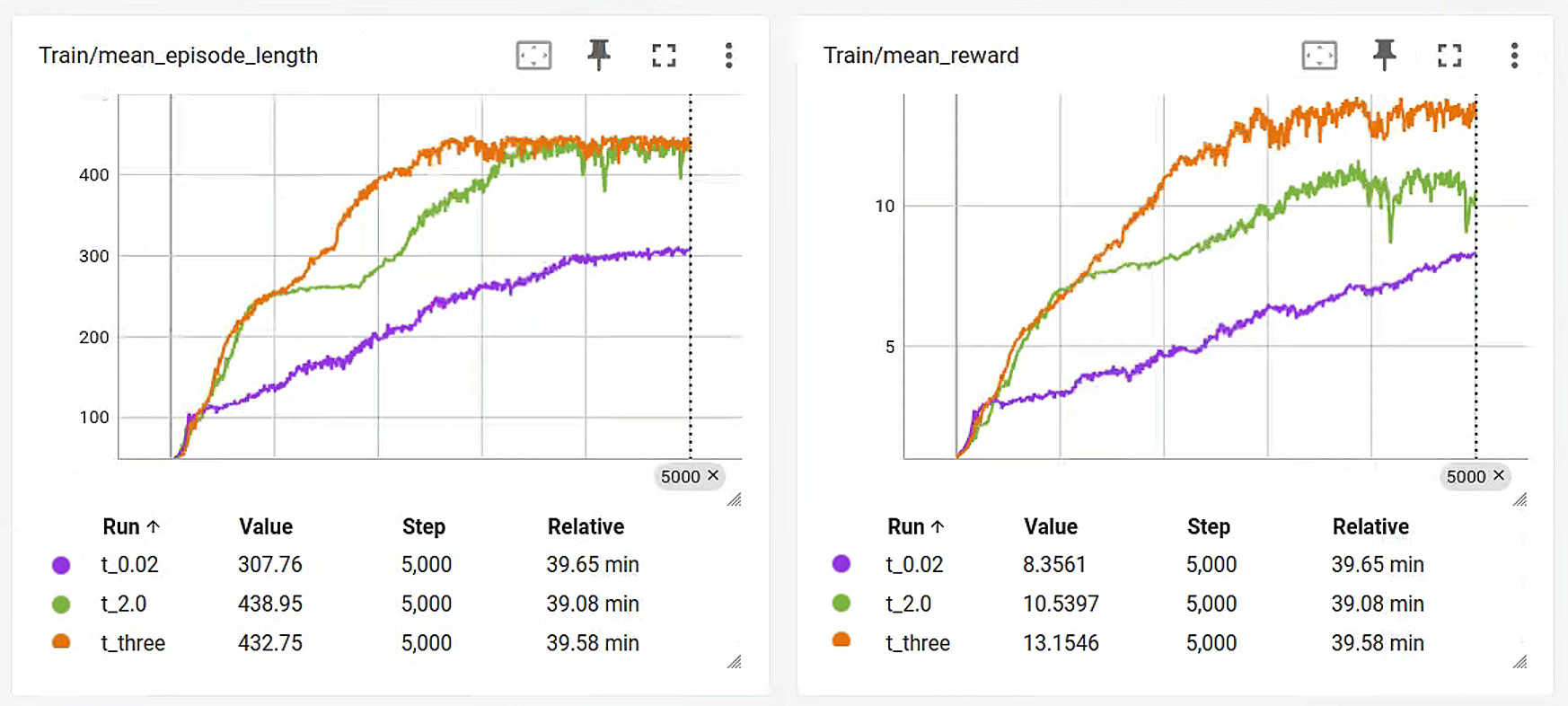}
    \caption{In the Genesis simulator, we compare single-temperature versus multi-temperature configurations on the high-difficulty dance motion tracking task. The results clearly demonstrate that the multi-temperature design is highly beneficial for training.}
    \label{fig:placeholder}
\end{figure}
 We further examine the Effective Sample Size (ESS) metrics as follows:
\begin{table}[H]
  \centering
  \caption{Multi-Temperature ESS Metrics}
  \label{tab:3x4}
  \begin{tabular}{|c|c|c|c|}
    \hline
    t & 0.02 & 0.15 & 2.0 \\
    \hline
    ESS\_ratio & \textbf{0.1123} & 0.4798 & 0.8612 \\
    \hline
    Max\_p & 0.8833 & 0.6711 & \textbf{0.1354} \\
    \hline
  \end{tabular}
\end{table}

\begin{itemize}
    \item For the single low-temperature case (\(t=0.02\)), the ESS ratio is only 0.1123. This extremely low value causes severe sample waste, which directly corresponds to the notably slow training progress observed in the figure. 
    \item For the single high-temperature case (\(t=2.0\)), Max\_p drops to 0.1354, indicating sufficient exploration. However, it comes at the cost of losing fine-grained reward details, resulting in consistently lower reward performance compared to the multi-temperature approach.
\end{itemize}

\begin{figure}[H]
    \centering
    \includegraphics[width=1.0\linewidth]{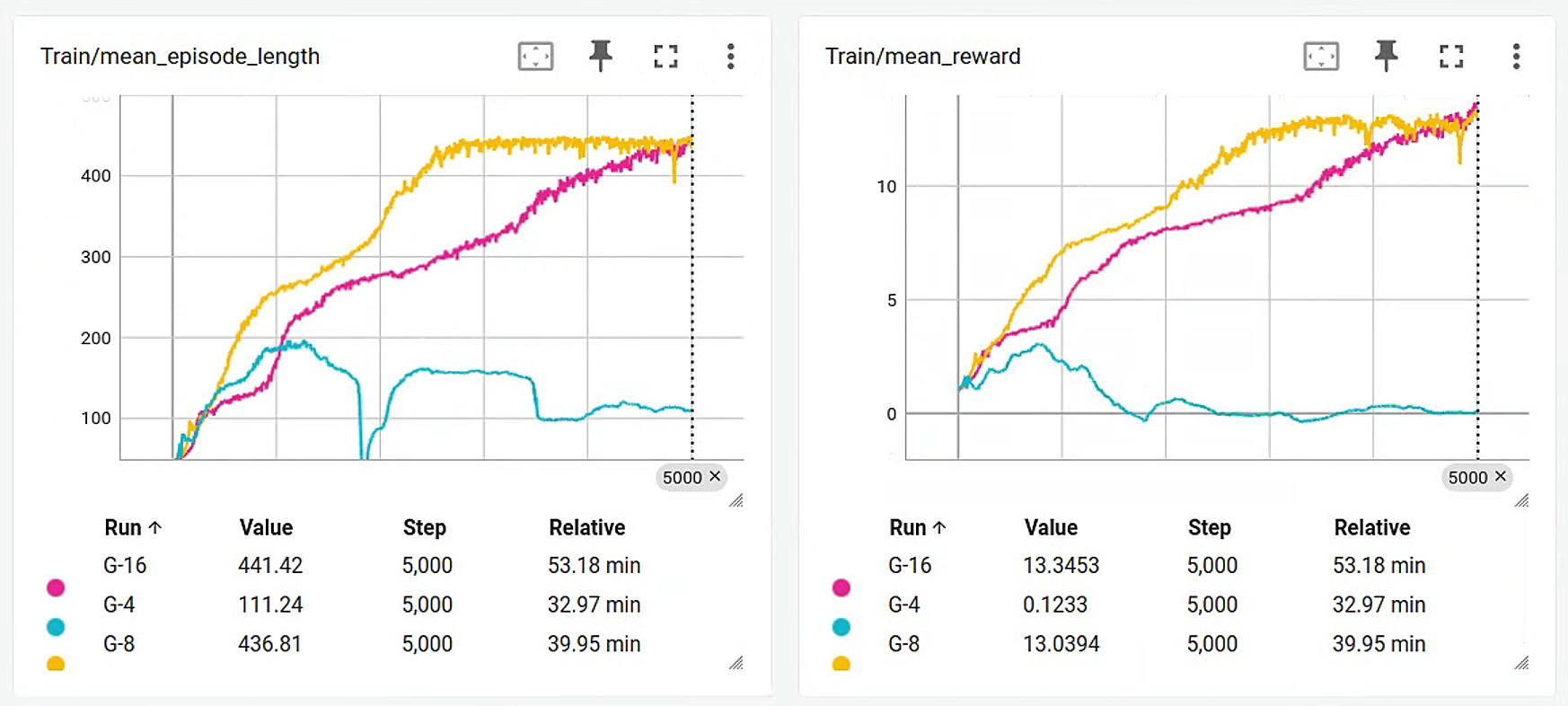}
    \caption{Ablation study on the number of candidate actions \(G\) for the high-difficulty dance motion tracking task in the Genesis simulator. Training curves for \(G=4\), \(G=8\) (default), and \(G=16\). When \(G=4\), the contrastive set per observation contains too few negative candidates, leading to highly unstable drifting vector estimation. The policy is easily pulled off track and collapses catastrophically to near-zero reward. \(G=16\) achieves strong final performance but trains slower than \(G=8\). In the same wall-clock time, increasing the number of parallel environments yields better overall results than simply increasing \(G\) to 16. Thus, we recommend \(G=8\) as the default choice.}
    \label{fig:placeholder}
\end{figure}

\section{Discussion}

Although PODPO demonstrates strong performance in robotic continuous control tasks, our experiments reveal noticeable reward jitter during the later stages of training. This jitter becomes progressively more pronounced as action diversity increases.

We attribute this phenomenon to a common characteristic of generative online reinforcement learning: the inherently multimodal nature of generative policies naturally gives rise to multiple local optima in the reward landscape across the action space, which in turn induces reward fluctuations. Preliminary tests indicate that this issue can be effectively mitigated by increasing the number of parallel environments or enlarging the training batch size.

In future work, we plan to further refine the constraint mechanisms on the drifting field to enhance the long-term stability of training.

\section{Summary}

The Positive-Only Drifting Policy Optimization (PODPO) proposed in this paper represents only an initial exploration in the field of generative online reinforcement learning. As an emerging direction that fuses generative models with reinforcement learning, generative RL still possesses tremendous untapped potential in robotic control. This potential spans a broad spectrum of exciting research opportunities, including further refinement of drifting-field constraint mechanisms, adaptive designs for multi-temperature strategies, deeper theoretical development of the positive-only learning paradigm, improved cross-task generalization, and enhanced robustness.

This work has validated the effectiveness of single-step drifting and positive-only updates specifically in legged robotic continuous control. Looking ahead, PODPO can be extended to more complex robotic scenarios, such as multi-robot collaboration, dynamic environment obstacle avoidance, and dexterous manipulation. We will also explore the integration of the positive-only optimization paradigm with other reinforcement learning frameworks, fully unleashing the intrinsic value of generative policies in efficient exploration, sample-efficient learning, and proactive error prevention. Ultimately, this will drive the stable, efficient, and large-scale deployment of generative reinforcement learning in real-world robotic systems.

\appendix

\section{Drifting Vector Computation (\texttt{compute\_V})}
The drifting vector \(V\) is computed as follows (see the original drifting models~\cite{podpo_drifting} for the general formulation). We adopt per-observation local contrastive computation with adaptive scaling and self-masking.

\begin{algorithm}[H]
\caption{compute\_V: Multi-Temperature Adaptive Contrastive Drifting Vector}
\label{alg:compute_v}
\begin{algorithmic}[1]
\REQUIRE Candidate actions \(x \in \mathbb{R}^{B \times G \times D}\), 
         positive actions \(y_{\rm pos} \in \mathbb{R}^{B \times N_{\rm pos} \times D}\), 
         negative actions \(y_{\rm neg} \in \mathbb{R}^{B \times M_{\rm neg} \times D}\),
         temperature set \(T = \{0.02, 0.15, 2.0\}\),
         self-masking flag \(\texttt{mask\_self} = \texttt{True}\)
\ENSURE Drifting vector \(V \in \mathbb{R}^{B \times G \times D}\)

\STATE Compute pairwise distances: 
       \(d_{\rm pos} \leftarrow \texttt{cdist}(x, y_{\rm pos})\), 
       \(d_{\rm neg} \leftarrow \texttt{cdist}(x, y_{\rm neg})\)

\IF{\(\texttt{mask\_self}\) and \(G = M_{\rm neg}\)}
    \STATE Mask self-repulsion: \(d_{\rm neg} \leftarrow d_{\rm neg} + 10^6 \cdot \mathbf{I}\)
\ENDIF

\STATE \(\text{all\_dists} \leftarrow \texttt{cat}([d_{\rm pos}, d_{\rm neg}], \dim=2)\)
\STATE \(\text{scale} \leftarrow \texttt{clamp}(\text{mean}(\text{all\_dists}[\text{all\_dists} < 10^5]), \min=10^{-3})\)

\STATE \(V \leftarrow \mathbf{0}_{B \times G \times D}\)

\FOR{\(\tau \in T\)}
    \STATE \(\text{logits} \leftarrow \texttt{cat}([-d_{\rm pos}/\tau, -d_{\rm neg}/\tau], \dim=2) / \text{scale}\)
    \STATE \(A \leftarrow \sqrt{\texttt{softmax}(\text{logits},\,2) \odot \texttt{softmax}(\text{logits},\,1)}\)
    \STATE Split: \(A_{\rm pos} \leftarrow A[:,:,:\,N_{\rm pos}]\), 
                  \(A_{\rm neg} \leftarrow A[:,:,N_{\rm pos}:]\)
    \STATE \(W_{\rm pos} \leftarrow A_{\rm pos} \odot A_{\rm neg}.\texttt{sum}(2,\texttt{keepdim})\)
    \STATE \(W_{\rm neg} \leftarrow A_{\rm neg} \odot A_{\rm pos}.\texttt{sum}(2,\texttt{keepdim})\)
    \STATE \(V_{\tau} \leftarrow W_{\rm pos} \cdot y_{\rm pos} - W_{\rm neg} \cdot y_{\rm neg}\)
    \STATE \(V \leftarrow V + V_{\tau}\)
\ENDFOR

\STATE \textbf{return} \(V\)
\end{algorithmic}
\end{algorithm}
\end{document}